\documentclass[letterpaper, 10 pt, conference]{ieeeconf}  % Comment this line out if you need a4paper

\IEEEoverridecommandlockouts                              % This command is only needed if 
\usepackage{url}
\usepackage{subfig}
\usepackage{graphicx}
\usepackage{varioref}
\graphicspath{{figures/}}
\labelformat{equation}{\textup{(#1)}}
\usepackage[pageanchor=true,plainpages=false,pdfpagelabels,bookmarks,bookmarksnumbered]{hyperref}
\usepackage{amssymb}
\usepackage{multirow}
\usepackage{todonotes}
\usepackage{flushend}

\usepackage{array}
\newcolumntype{L}[1]{>{\raggedright\let\newline\\\arraybackslash\hspace{0pt}}m{#1}}
\newcolumntype{C}[1]{>{\centering\let\newline\\\arraybackslash\hspace{0pt}}m{#1}}
\newcolumntype{R}[1]{>{\raggedleft\let\newline\\\arraybackslash\hspace{0pt}}m{#1}}

\def\usenatbib{1}
\ifx\usenatbib\undefined
    \usepackage{cite}
\else
    \makeatletter
    \let\NAT@parse\undefined
    \makeatother
    \usepackage[numbers,sort]{natbib}

    \setcitestyle{citesep={], [}}
    \makeatletter
    \def\NAT@def@citea{\def\@citea{\NAT@separator}}%
    \makeatother
\fi

\graphicspath{ {./figures/} }

\let\orgautoref\autoref
\providecommand{\Autoref}
        {\def\equationautorefname{Equation}%
         \def\figureautorefname{Figure}%
         \def\subfigureautorefname{Figure}%
         \def\Itemautorefname{Item}%
         \def\tableautorefname{Table}%
         \def\exerciseautorefname{Exercise}%
         \def\starexerciseautorefname{Exercise}%
         \def\sectionautorefname{Section}%
         \def\subsectionautorefname{Section}%
         \def\subsubsectionautorefname{Section}%
         \def\chapterautorefname{Section}%
         \def\partautorefname{Part}%
         \orgautoref}

\renewcommand{\autoref}
        {\def\equationautorefname{Equation}%
         \def\figureautorefname{Fig.}%
         \def\subfigureautorefname{Fig.}%
         \def\Itemautorefname{item}%
         \def\tableautorefname{Table}%
         \def\exerciseautorefname{Exercise}%
         \def\starexerciseautorefname{Exercise}%
         \def\sectionautorefname{Section}%
         \def\subsectionautorefname{Section}%
         \def\subsubsectionautorefname{Section}%
         \def\chapterautorefname{Section}%
         \def\partautorefname{Part}%
         \orgautoref}

\usepackage{hyperref}
\hypersetup{
colorlinks=true,
urlcolor=blue,
}
\title{\LARGE \bf
Self-supervised Representation Learning for \\
Reliable Robotic Monitoring of Fruit Anomalies
}

\author{Taeyeong Choi$^{1}$, Owen Would$^{2}$, Adrian Salazar-Gomez$^{1}$, 
        and Grzegorz Cielniak$^{1}$% <-this % stops a space
\thanks{All authors are with the Lincoln \mbox{Agri-Robotics (LAR) Centre, University} of Lincoln, UK. $^{1}${\tt\small \{tchoi, asalazargomez, gcielniak\} @lincoln.ac.uk}, $^{2}${\tt\small 25393497@students.lincoln.ac.uk}
        }%
}

\begin{document}

% Copyright notice for update to open access repositories after accepted
\begin{figure*}
\begin{minipage}{1.\textwidth}
        {\textcopyright } 2022 IEEE.  Personal use of this material is permitted.  
        Permission from IEEE must be obtained for all other uses, in any current or future media, 
        including reprinting/republishing this material for advertising or promotional purposes, 
        creating new collective works, for resale or redistribution to servers or lists, or reuse 
        of any copyrighted component of this work in other works.
\end{minipage}
\end{figure*}

\pagebreak
%%%%%%%%%%%%%%%%%%%%%%%%%%%%%%%%%%%%%%%%%%%%%%%%%%%%%%%%%%%%%%%%%%%%%%%%%%%%%%%%

\maketitle
\thispagestyle{empty}
\pagestyle{empty}

%%%%%%%%%%%%%%%%%%%%%%%%%%%%%%%%%%%%%%%%%%%%%%%%%%%%%%%%%%%%%%%%%%%%%%%%%%%%%%%%
\begin{abstract}

% Data augmentation can be a simple yet powerful tool for autonomous robots 
% to fully utilise available data to perform self-supervised learning of identifying
% \emph{atypical} scenes or objects from their past experiences.
% In computer vision communities, successful augmentation methods have been implemented by 
% arbitrarily embedding structural peculiarity in focal objects on typical images 
% so that classifying these artefacts can provide guidance of learning useful 
% representations for the detection of anomalous visual inputs.
Data augmentation can be a simple yet powerful tool for autonomous robots 
to fully utilise available data for self-supervised identification of \emph{atypical} scenes or objects. 
State-of-the-art augmentation methods arbitrarily embed ``structural'' peculiarity 
on typical images so that classifying these artefacts can provide guidance for learning representations for the detection of 
anomalous visual signals.
In this paper, however, we argue that learning such structure-sensitive representations 
can be a suboptimal approach to some classes of anomaly
(e.g.,~unhealthy fruits) which could be better recognised by a different type of 
visual element such as ``colour''.
We thus propose \emph{Channel Randomisation} as a novel data augmentation 
method for restricting neural networks to learn encoding of ``colour irregularity'' 
whilst predicting \emph{channel-randomised} images to ultimately build reliable 
fruit-monitoring robots identifying atypical fruit qualities.
% \todo{Can we strengthen this argument here? e.g. structure-sensitive representations are not suitable for all classes of visual anomalies, 
% and appearance-based alternatives such as colour are better suited for specific domains such as agriculture.}
% \todo{Abstract: I would swap the contributions presented so first the method and then the dataset.} 
Our experiments show that (1)~this colour-based alternative can better learn 
representations for consistently accurate identification of fruit anomalies 
in various fruit species, and also,  
(2)~unlike other methods, the validation accuracy can be utilised as a criterion 
for early stopping of training in practice due to positive correlation between the 
performance in the self-supervised colour-differentiation task and 
the subsequent detection rate of actual anomalous fruits.
Also, the proposed approach is evaluated on a new agricultural dataset, 
\emph{Riseholme-2021}, consisting of $3.5$K~strawberry images gathered by a 
mobile robot, which we share online to encourage active agri-robotics 
research.  

\end{abstract}
% \todo{The abstract can be made a bit shorter. Here are some proposals: Data augmentation can be a simple yet powerful tool for autonomous robots 
% to fully utilise available data for self-supervised identification of \emph{atypical} scenes or objects. State-of-the-art augmentation methods arbitrarily embed structural peculiarity of focal objects in typical images so that classifying these artefacts can provide guidance for learning representations for the detection of anomalous visual inputs.

% In addition, the proposed approach is evaluated on a new anomaly dataset \emph{Riseholme-2021}, consisting of $3.5$K~strawberry images collected in-field from a mobile robot, which we share with the community to encourage active agri-robotics research.  
% }

%%%%%%%%%%%%%%%%%%%%%%%%%%%%%%%%%%%%%%%%%%%%%%%%%%%%%%%%%%%%%%%%%%%%%%%%%%%%%%%%
\section{INTRODUCTION}
\label{sec:introduction}

% \todo{Perhaps this opening is a bit more concise: Agricultural mobile robots are expected to precisely assess the qualities of crops from their sensory information to autonomously perform the targeted treatment of individual plants or harvest the mature and healthy crops. To realise this autonomy, deep learning models could be adopted to classify visual input from robotic sensors by optimising their parameters based on a large number of examples available in advance.}

% In agri-robotics, mobile robots are expected to be able to precisely assess the 
% qualities of crops from the perceived information to autonomously perform 
% the best treatments specialised for each individual or harvest only the ones 
% that are healthy and sufficiently mature.
% To realise this autonomy, deep learning models could be adopted to classify 
% graphical inputs from the robotic visual sensors by optimising their parameters
% based on a large number of examples available in advance. 
Agricultural mobile robots are expected to precisely assess the qualities of crops from their sensory information to autonomously perform the targeted treatment of individual plants or harvest the mature and healthy crops. To realise this autonomy, deep learning models could be adopted to classify visual input from robotic sensors by optimising their parameters based on a large number of examples available in advance.
In practice, however, collecting data of ``atypical'' qualities,
e.g., fruits with disease or damage, can be challenging mainly because of their rare occurrences, and 
therefore, One-class Classification~(OC) paradigm~\citep{CPLP21,RVGDSBMK18} 
has been widely used in computer vision communities, in which classifiers are 
trained to maximise the utility of the available data from ``normal'' class 
to later distinguish unseen instances of ``anomalous'' class as well.

Self-supervised Learning~(SL) has been introduced as a powerful method to effectively 
solve OC~problems by augmenting training data to inject some level of 
\emph{unusual} patterns because classifying the artefacts can be an instructive proxy task 
to learn potentially informative feature representations for detecting anomalies in 
tests~\citep{LSYP21,DT17,YY20,HMKS19}. 
Nonetheless, most successful SL~tasks have been designed only for the scenarios in which 
anomaly is mostly defined by \emph{structural} differences---e.g.,~bent tips in screws, 
holes on hazelnut bodies, or missing wires in cable clusters in MVTec AD 
Dataset~\citep{BBFSS21}---or image samples out of a particular training class in 
large datasets such as ImageNet~\citep{DDSLLF09} or CIFAR-$10$~\citep{KH09}. 
\begin{figure}[t]\centering
        \subfloat[]{\label{fig:thorvald}%
        \includegraphics[width=.95\linewidth]{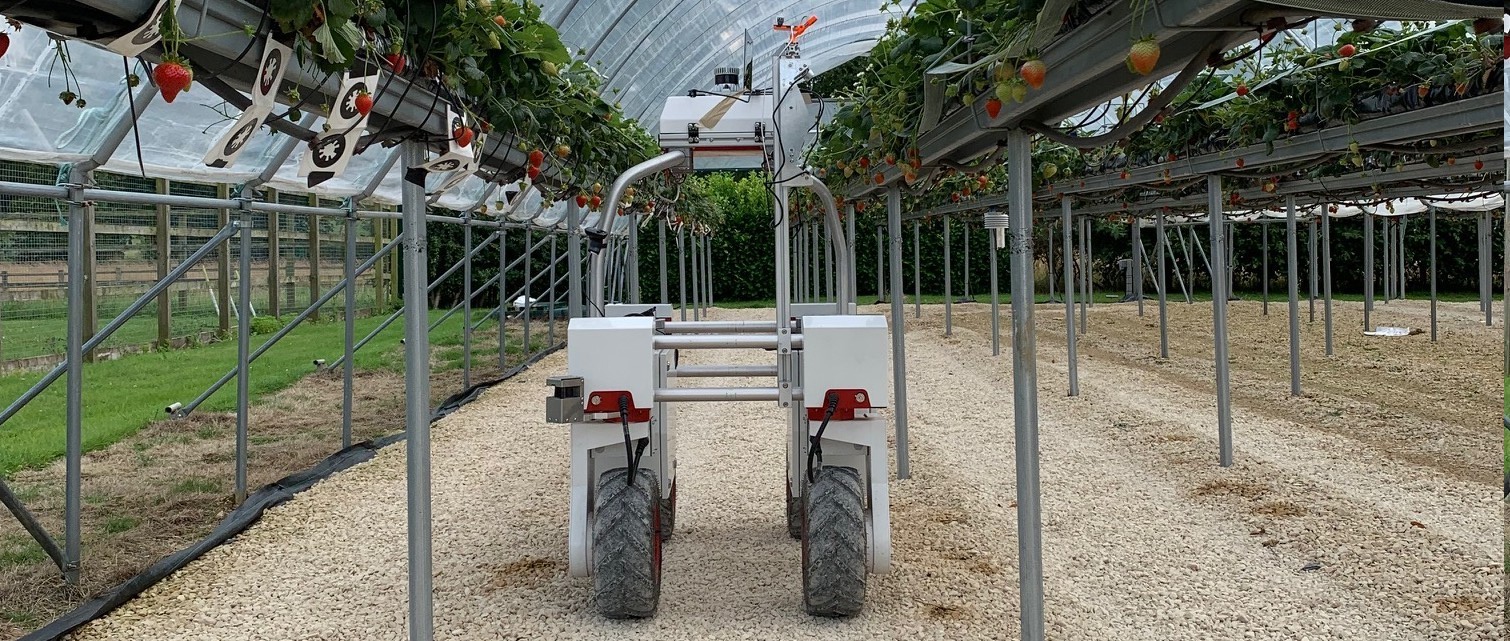}}\\[-2ex]
        \subfloat[]{\label{fig:strawberry_examples_normal}%
        \includegraphics[width=.55\linewidth]{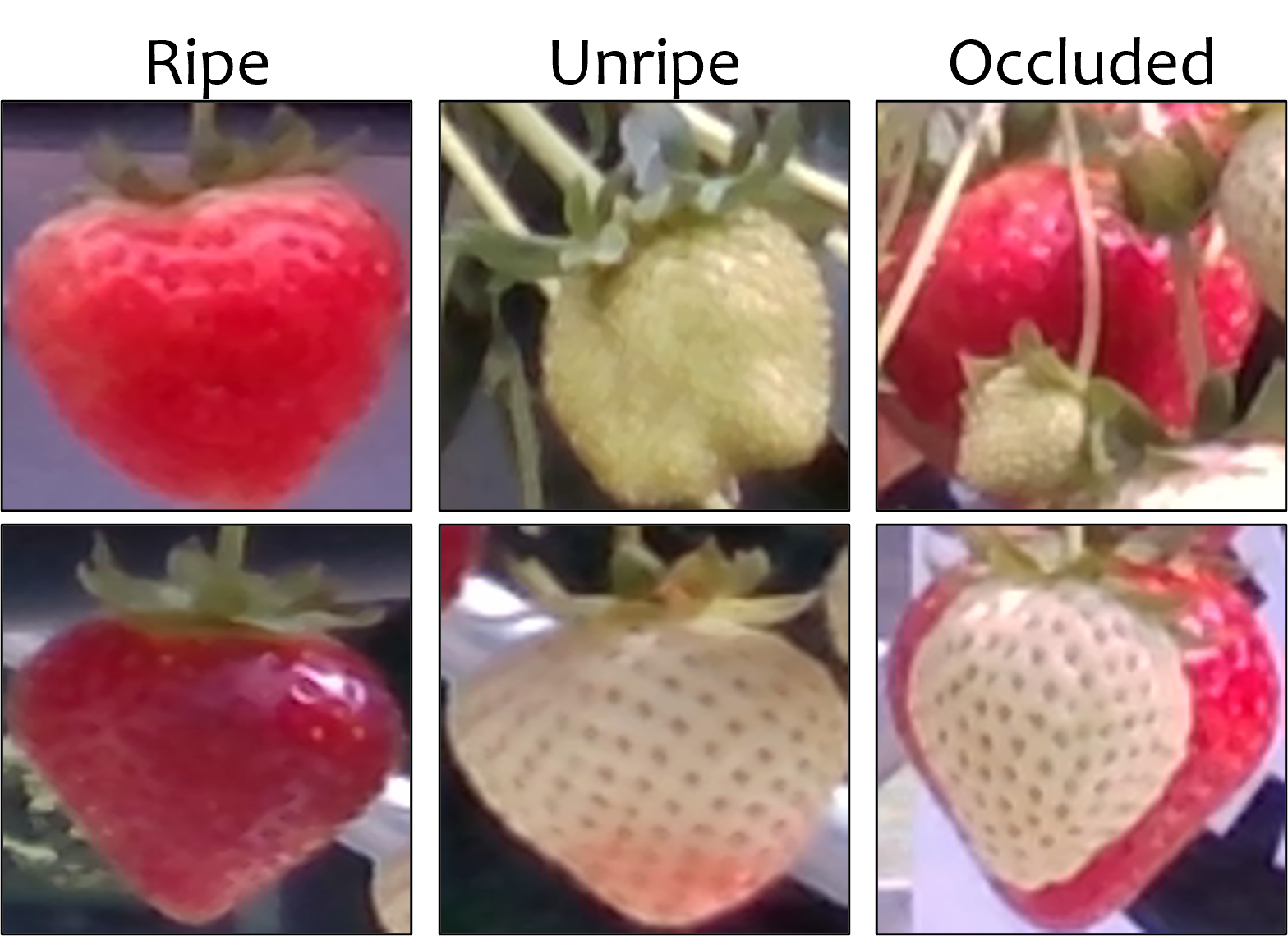}}
        \quad
        \subfloat[]{\label{fig:strawberry_examples_anomalous}%
        \includegraphics[width=.363\linewidth]{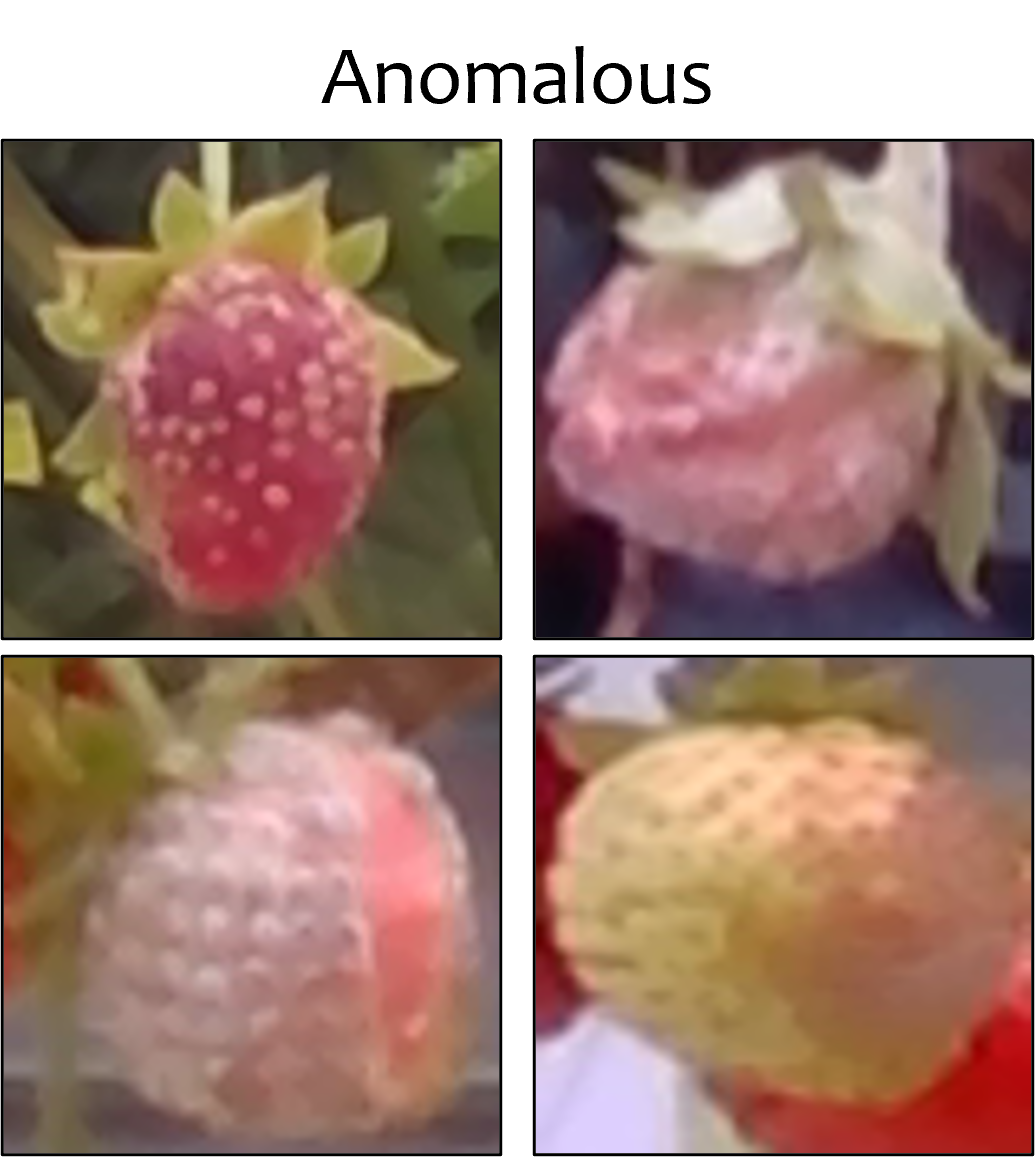}}
            \caption{\protect\subref{fig:thorvald}~Robotic monitoring of anomalous 
                strawberries using a mobile sensing platform \emph{Thorvald}; 
                Sample categories from Riseholme-2021 dataset including
                \protect\subref{fig:strawberry_examples_normal}~three ``normal'' subcategories
                and \protect\subref{fig:strawberry_examples_anomalous}~``anomalous''.}
            \label{fig:strawberry_examples}
\end{figure}
% \todo{I would change the caption to: Robotic monitoring of anomalous strawberries using a mobile sensing platform Thorvald (top). 
% Sample categories from Riseholme-2021 dataset including three ``normal'' categories (bottom-left) and ``anomalous'' (bottom-right).}

We argue that such representation learning techniques
may only provide suboptimal performance for~OC in agricultural 
domains since anomalies in fruits, for example, tend to appear with only 
little distinction in shape, but peculiar \emph{pigmentations}
(e.g.,~\autoref{fig:strawberry_examples_anomalous}) instead could be more 
useful visual cues for differentiation.
As an alternative, in this paper, we thus propose 
\emph{Channel Randomisation}~(CH-Rand), 
which augments each image of normal fruit by randomly permutating RGB~channels with 
a possibility of repetition to produce \emph{unnatural} ``colour'' compositions 
in the augmented image.  
Whilst classifying these artefacts, the neural networks automatically learn 
discriminative representations of irregular colour patterns, so that distance-based 
heuristics can later be employed on that learnt space to estimate the anomaly score of 
input using the distance to the existing data points.
% \todo{Add that this is not merely the feature space expansion.}

% \todo{Perhaps better realistic scenario rather than realistic performance: e.g. To validate the performance of our system in a realistic scenario...}
To validate the performance of our system in a realistic scenario, we also introduce 
\emph{Riseholme-2021}, a new dataset of strawberry images, for which we 
operated a mobile robot (\autoref{fig:thorvald}) to collect $3,520$~images of healthy and unhealthy 
strawberries at three unique developmental stages with possible occlusions
(cf.~\autoref{fig:strawberry_examples_normal}-\ref{fig:strawberry_examples_anomalous}). 
Our experiments are conducted not only on this set of strawberry images
but also on \emph{Fresh \& Stale} dataset with several other fruits
to show that CH-Rand can gain the most reliable representations 
for detection of anomalous fruits compared to all other baselines 
including self-supervised structure-learning methods (e.g.,~CutPaste~\citep{LSYP21}).
% in anomaly detection.
We further support our design by demonstrating high degrees of correlation 
between the success in colour prediction task and the final performance in anomaly 
identification.
%  compared to structure-based learning task. 
% \todo{would structure-based learning task be more specific?}
Hence, CH-Rand does not require manually engineered criteria for early stopping, 
and validation accuracy can simply be monitored during the proxy task to ensure 
the precise detection of actual anomalies.

\section{RELATED WORK}
\label{sec:related_work}

\subsection{Anomaly Detection in Agricultural Domains}
\label{sec:ad_in_agri}

Perception models have played a crucial role also in 
agriculture to build up essential capabilities to eventually 
deploy fully autonomous robots in real farms. 
For instance, weeds are targeted anomalies to detect in~\citep{EMAVF21, WALSP20, BADC20}, 
and occlusions or dense fruit clusters are of interest in~~\citep{MHG20, GXF20}.
% weeds can be a 
% target anomaly to spatially localise so that mobile robots could 
% selectively apply herbicide only to the detected regions. 
% identifying formational anomalies such as occlusion 
% or dense fruit clusters can also be considered to operate 
% robotic pickers without collisions which might damage the fruit~\citep{MHG20, GXF20}. 
More relevantly to our work, plant diseases are also important anomalies 
to detect as in~\citep{F18, SCJJ19, GWSBLA21}, in which 
networks were trained with annotated images to learn leaves with diseases. 
% Similarly, \citet{GWSBLA21} applied 
% segmentation algorithms to spot late blights on potato plant images.
These methods were built upon supervised learning aided
by manually annotated data, but our approach is designed to
meet the practical assumption in~OC that anomalous data may be 
unavailable during training. 
\citet{HHR19} also utilised colour-based features to recognise anomalous 
leaves, but they only depended on 
human engineered features, while ours trains deep neural networks.  
% to manually compute colour-based features to recognise anomalous leaves, 
% but in contrast, we propose a method to guide deep neural networks to automatically 
% discover useful features from augmented images with randomised 
% colour channels.

% \subsection{Self-supervised Learning for One-class Classification}
\subsection{One-class Classification Strategies}
\label{sec:oc_strategies}

% Due to unavailability of anomalous data during training, 
Due to the strict assumption in OC,
generative model-based frameworks have been widely used. 
% to tackle the unique challenges in~OC. 
For instance, Deep Convolutional Autoencoders~(DCAE) measure 
the reconstruction error because novel data would more likely 
cause higher errors~\citep{CPLP21,KWWBKA19}.
With DCAE as a backbone, \citet{RVGDSBMK18} introduced Deep Support 
Vector Data Decription~(DSVDD) to learn as dense 
representations near a central vector~$\vec{c}$ as possible so that atypical 
data points would be detected by a long distance from it. 
Generative Adversarial Networks~(GANs) can also provide a large benefit 
by synthesis of data to potentially model unavailable 
anomalous samples~\citep{CPLP21, SKFA18, SSWLS19, PNX19}. 
% Key concepts from DCAE and DSVDD are still used in GAN-based models. 
% For example, the generator in~\citep{SKFA18} is a form of 
% DCAE, and 
For example, IO-GEN~\citep{CPLP21} utilises a trained DSVDD to
replace its $\vec{c}$ with synthetic data to perform multi-dimensional 
classification for complex datasets in place of
simplistic distance calculation.

\subsubsection{Self-supervised Learning}
\label{sec:sl}

% Similar to GANs, SL~is configured to produce artificial data, but they are 
% largely the augmented products by relatively simple manipulations 
% on training data.  
The ultimate goal in SL is to gain useful representations in neural 
networks for future anomaly detection 
whilst identifying intentionally manipulated data as a \emph{pretext} task. 
For example, inferring (1)~geometric transformations such as rotation~(ROT) 
applied to input images~\citep{GSK18},
(2)~relative locations among regional patches~\citep{YY20}, 
or (3)~images with blank local masks embedded~\citep{DT17} 
has shown great successes for~OC.
% \todo{it is clear that CIFAR is a dataset with industrial anomalies?}  
More recently, \citet{LSYP21} introduced CutPaste~(CP),
in which unlike~\citep{DT17}, local patches are extracted directly from the 
original images (cf.,~\autoref{fig:cutpaste}) to keep 
the pretext task more challenging.
% each image is augmented with a local image patch in it. 
% Unlike in~\citep{DT17}, the local patch is extracted from 
% the augmented image to keep the proxy task more challenging
% (cf.,~\autoref{fig:cutpaste}). 

% As clearly mentioned in~\citep{LSYP21}, 
In fact, all these augmentations were motivated to model typical 
structures of normal objects (e.g.,~defect-free screws) to detect odd shapes 
in anomalous examples (e.g.,~screws with bent tips) afterwards.
We, however, argue that such structural difference may be less significant 
in differentiation between healthy and unhealthy fruits, and we 
propose to learn \emph{colour regularity} instead as an alternative.
%  could provide more instructive guidance to 
% obtain useful representation. 

\subsection{Channel Randomisation for SL}
\label{sec:channel_randomisation_for_sl}
% \todo{Perhaps this can be just a paragraph without a section as it logically is connected with the previous subsection.}

Although colourisation can be used to learn useful representations by colouring 
grayscale images~\citep{VSFGM18, LMS17}, 
a more relevant technique to CH-Rand is \emph{Channel Permutation} 
(CH-Perm)~\citep{LHS20}, also called \emph{Channel Swap}~\citep{LHSY17}, 
in which five random permutations---i.e.,~RBG, GRB, GBR, BRG, and BGR---can be 
considered for augmentation by reordering the channels of a RGB~image without 
repetition.
% Image classifiers took some benefits from predicting the transformations, 
% but less improvement was observed than other geometric augmentation methods
% such as rotation in~\citep{LHS20}. 
\emph{Channel Splitting} (CH-Split)~\citep{LHSY17} is also related,
in which the values of a randomly chosen channel---R, G, or B---are all 
copied to others to potentially produce three novel visuals.
\citet{LHSY17} applied this to their SL~framework to encourage their 
models to ``ignore'' colour variations but learn semantic coherence 
through a human action. 
Contrary to any of these techniques, CH-Rand here is adopted in the context of 
one-class classification to ``reflect'' colour regularities in representation 
learning. 
Moreover, CH-Rand can generate a larger set of $26$~random channel 
sequences including the ones that CH-Perm and CH-Split can generate, so  
in~\autoref{sec:experiments}, we investigate the benefits from using 
it.

\section{METHODOLOGY}
\label{sec:methodology}

As in previous approaches~\citep{LSYP21}, our framework includes two 
modular processes for anomaly identification on fruit images: 
(1)~self-supervised representation learning with data augmentation and
(2)~anomaly score estimation. 
In~\ref{sec:ch_rand}, we first formalise our proposed CH-Rand augmentation
and then in~\ref{sec:decision_of_anomaly}, describe a heuristic method 
to calculate anomaly scores on learnt representations. 
\begin{figure}[t]
        \centering
        \subfloat[]{\label{fig:cutpaste}%
        \includegraphics[width=.19\linewidth]{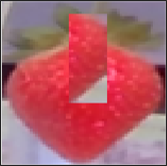}}
        \subfloat[]{\label{fig:patch_brr}%
        \includegraphics[width=.19\linewidth]{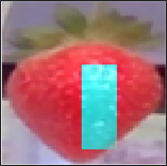}} 
        \subfloat[]{\label{fig:threshold_brr}%
        \includegraphics[width=.19\linewidth]{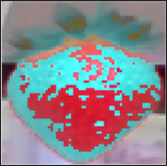}}
        \subfloat[]{\label{fig:sobel_brr}%
        \includegraphics[width=.19\linewidth]{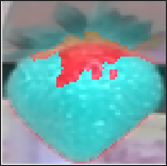}}
        \subfloat[]{\label{fig:entire_brr}%
        \includegraphics[width=.19\linewidth]{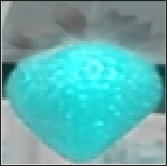}}
        \caption
        {
        Various data augmentations on an image of ripe strawberry: 
        \protect\subref{fig:cutpaste} CutPaste~\citep{LSYP21}, and 
        the permutation of ``BRR'' applied
        to \protect\subref{fig:patch_brr}~a local patch, 
        \protect\subref{fig:threshold_brr}~$50\%$~thresholded pixels, 
        \protect\subref{fig:sobel_brr}~a segment of 
        \emph{Sobel} filter~\citep{scikit}, 
        and \protect\subref{fig:entire_brr}~all pixels.
        }
        \label{fig:augmentation_examples}
\end{figure}

\subsection{Channel Randomisation}
\label{sec:ch_rand}

Our approach is motivated by the unique observations of fruit anomalies
compared to other types---e.g., local defects on industrial products 
in MVTec AD~\citep{BBFSS21} or the out-of-distribution samples in 
CIFAR-$10$~\citep{KH09}. 
To be specific, as shown in~\autoref{fig:strawberry_examples_normal},
fruits generally have relatively high phenotypic variations
in local structures even in the same species regardless of normality; 
% \todo{do you mean here freshness or ripeness?}
nevertheless, healthy fruits at the same developmental stage all share  
a similar colour composition, which can change dramatically as the fruit 
becomes unhealthy, for example due to some fungal infection as displayed 
in~\autoref{fig:strawberry_examples_anomalous}. 
Therefore, we design a novel augmentation method to restrict the 
neural network to learn representations for encoding 
\emph{colour irregularity} to ultimately build a more reliable 
anomaly detector on agricultural robots.
% \todo{I like this sentence: I would repeat that in the abstract and conclusions.}

CH-Rand can be simply performed by computing a random permutation 
of colour channels with a possibility of repetition to apply to the entire image input. More formally, we generate an augmented image~$\mathcal{A} \in \mathbb{R}^{W\times H\times C}$
by executing CH-Rand on the original image~$\mathcal{I} \in \mathbb{R}^{W\times H\times C}$
of normal class available during training, where $W$ and $H$ are the width and the
height, respectively, and $C$~is the number of channels, which is typically
set to $3$~for the RGB image format. 
% For each spatial element~$a^{c}_{w,h}$ in $\mathcal{A}$, 
To augment a new input~$\mathcal{I}$, we first randomly 
build an arbitrary function $\pi: \chi \rightarrow \chi'$ for permutation,
where $\chi = \{1,2,...,C\}$ denotes the indices of original channels, 
and $\chi' \in \mathcal{P}(\chi) \setminus \varnothing$
as $\mathcal{P}$~returns the powerset of input.
Each element~${a}_{w,h}^{c}$ in $\mathcal{A}$ can then be 
determined as follows: 
\begin{equation}
        {a}_{w,h}^{c} = i_{w,h}^{\pi{(c)}},
        \label{eq:augmentation}
\end{equation}
for which $\pi$~is \emph{fixed} for every $w,h,$ and $c$ to apply the same 
channel assignment during a single augmentation process.

Note here that the output sequence of $\pi$ may use some \emph{duplicate} 
channel indices from~$\chi$ by design because $|\chi'| \leq |\chi|$. 
Moreover, we keep drawing a new $\pi$ until 
$\exists c\in\chi, c\neq\pi(c)$ to avoid the case of 
$\mathcal{A}=\mathcal{I}$. 
Consequently, 
$26$~possible channel sequences exist for augmentation in $3$-channel format, 
whereas CH-Split and CH-Perm only have three and five~possibilities, 
respectively.
An example of augmentation is presented in~\autoref{fig:entire_brr}, for which the channel sequence of BBR has been generated from the RGB colour space.
% \todo{There is something funny with \O above. what about $\varnothing$?}

Based upon this augmentation method, a classifier can be set to 
learn the binary classification to predict whether input images
are the products of the augmentations. 
Inspired by~\citep{LSYP21, GSK18}, we design our loss function below 
to train a deep neural network-based classifier~$f_\Theta$
on a training dataset~$\mathcal{D}$:
\begin{equation}
        \mathcal{L} = 
        \mathbb{E}_{\mathcal{I}\in\mathcal{D}} \Big[ 
        H(f_{\Theta}(\mathcal{I}), 0) + H(f_{\Theta}(CHR(\mathcal{I})), 1) \Big],
        \label{eq:loss}
\end{equation}
where $CHR$~is the application of CH-Rand augmentation, and 
$H$~is the function of binary cross entropy to 
estimate the prediction error in classification. 
In implementation, we randomly sample a 
batch~$\mathcal{D}' \subseteq \mathcal{D}$ at 
each iteration to feed a half with augmentation and the other without.

\subsection{Anomaly Detection}
\label{sec:decision_of_anomaly}
% \todo{maybe Anomaly Detection is a more straightforward term?}
For anomaly prediction, we use the feature representations~$g_\theta$ learnt 
within the classifier~$f_\Theta$---i.e.,~$g_\theta$ is the output of an 
intermediate layer in~$f_\Theta$.
%  -- whilst performing the binary classifications 
% described in~\autoref{sec:sl4oc}. 
% Technically, $g_\theta$ can be obtained from the output of an intermediate
% layer in~$f_\Theta$. 
Whilst minimising the loss function in~\autoref{eq:loss}, the representations
of the normal training data~$\mathcal{D}$ are likely to be clustered 
to maximise the \emph{distance} from anomalous data to effectively separate them. 
Therefore, similar to~\citep{PP19}, we calculate the anomaly score~$s$ for 
an input image~$\mathcal{I}'$ by computing the average distance to the 
$k$~nearest neighbors~$\mathcal{N}\subseteq\mathcal{D}$ 
in the space of $g_\theta$:
$s(\mathcal{I}') = (1/k)\sum_{\mathcal{I}\in\mathcal{N}} 
\delta \big( g_\theta(\mathcal{I}), g_\theta(\mathcal{I}') \big)$, 
where $\delta$~returns the Euclidean distance between two input vectors.

Our design is generic to easily replace this scoring module 
with other unsupervised techniques, such as Gaussian density 
estimators~\citep{RMM21} or One-class SVM~\citep{SPSSW01}.
Yet, we use the $k$-neighbour heuristic since it has performed 
best in our tests. 

\section{EXPERIMENTS}
\label{sec:experiments}
\begin{figure}[t]
        \centering
        \subfloat[]{\label{fig:rep_ripe}%
        \includegraphics[width=.33\linewidth]{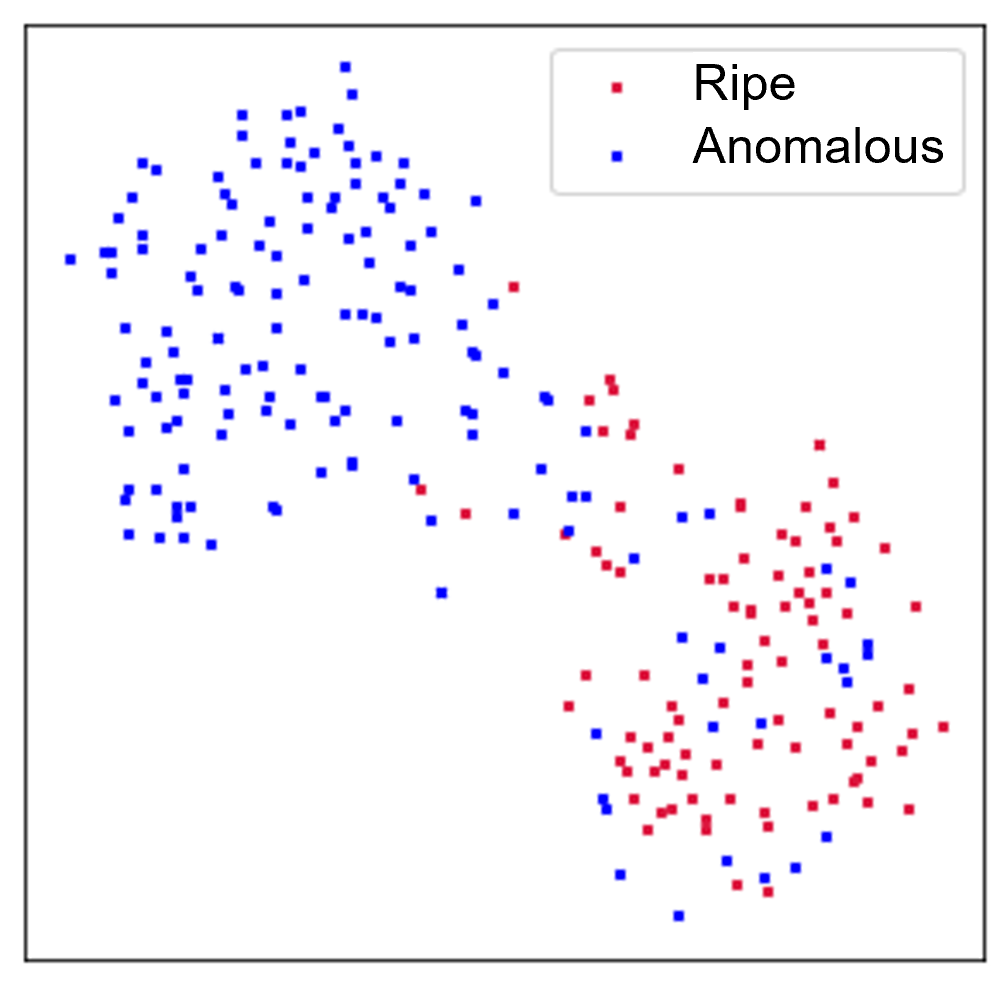}}
        \subfloat[]{\label{fig:rep_unripe}%
        \includegraphics[width=.33\linewidth]{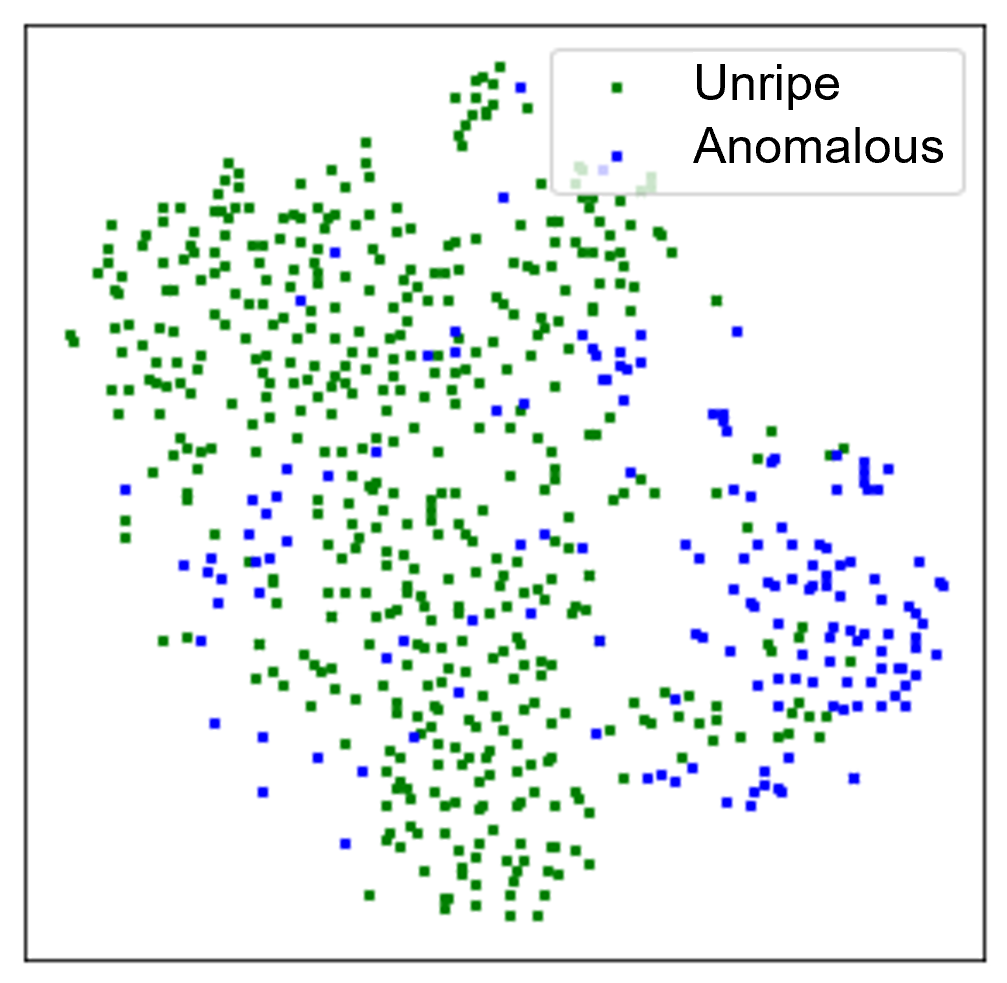}} 
        \subfloat[]{\label{fig:rep_ripe_unripe}%
        \includegraphics[width=.33\linewidth]{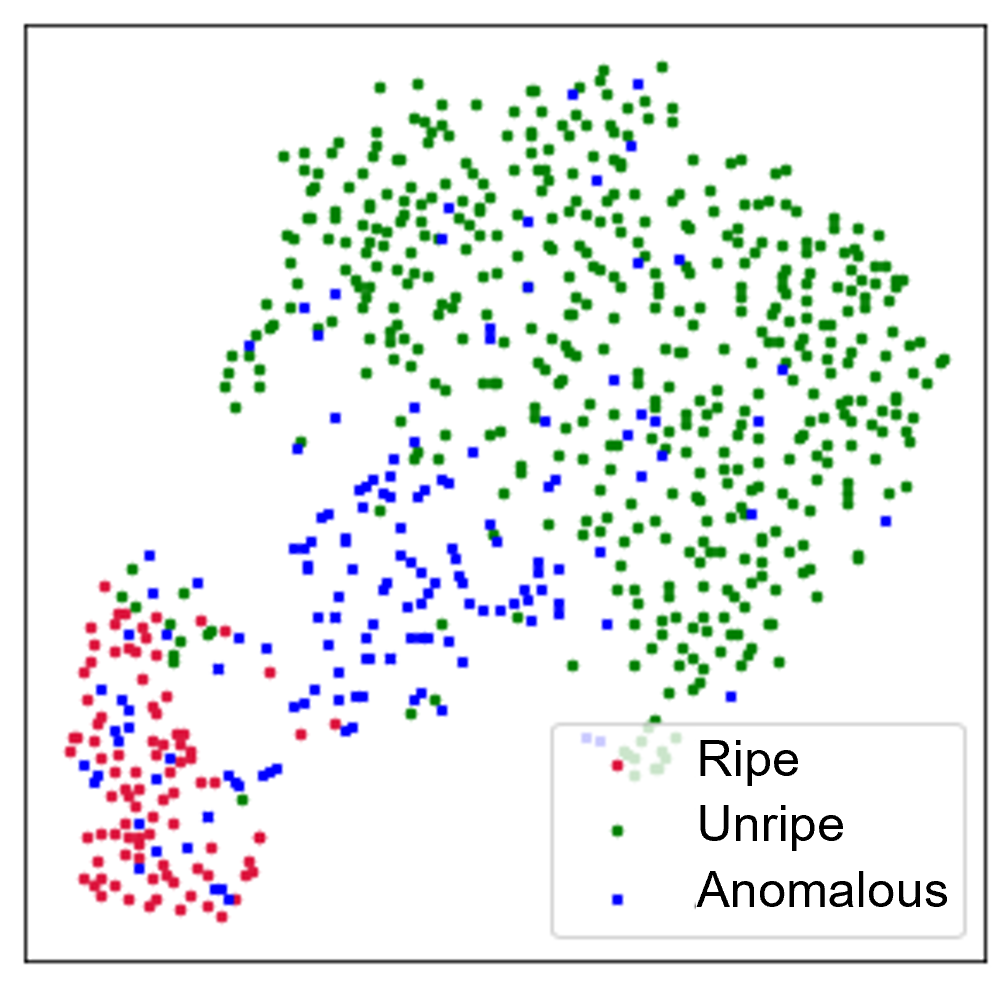}}
        \caption
        {
        t-SNE visualisation of learnt representations with~CH-Rand
        on Riseholme-2021 as the normal class for training only included
        \protect\subref{fig:rep_ripe}~Ripe,
        \protect\subref{fig:rep_unripe}~Unripe, or
        \protect\subref{fig:rep_ripe_unripe}~Ripe\&Unripe.
        }
        \label{fig:rep_viz}
\end{figure}

% \todo{If you need to cut down some text the following paragraph is the first candidate. Perhaps that could be shortened to 2-3 sentences.}

We here offer the experimental results to demonstrate 
the performance of our proposed framework in vision-based 
monitoring of fruit anomalies. We first explain the fruit-image datasets for 
anomaly detection and technical details used through experiments 
in~\autoref{sec:datasets} and~\autoref{sec:implementation_and_protocols}, 
respectively. 
\Autoref{sec:comparative_results} then shows quantitative results by 
comparison with other baselines, and based on the results, we examine 
in~\autoref{sec:relevance_of_sl_task} the relevance between the pretext
task that CH-Rand generates and the subsequent detection of 
real unhealthy fruits.
Lastly, \autoref{sec:ablation_study} includes ablation studies to discuss 
variants of~CH-Rand. 

% In~\autoref{sec:implementation_and_protocols}, we first describe 
% the technical details of our implementations as well as the 
% evaluation schemes. 
% Furthermore, we perform ablation studies in~\autoref{sec:ablation_study} to 
% understand  ``where'' and ``which'' randomisations should apply to 
% gain the best results,  and the considerations on the optimal input size and 
% the choice of intermediate layer also follow. 
% \autoref{sec:comparative_results} then shows quantitative comparisons with
% baseline models on two fruit image datasets -- 
% \emph{Riseholme-2021} and   \emph{Fresh \& Stale} -- to evaluate 
% applicability to various types of fruit. 
% We also investigate the relevance of the proposed proxy task for SL
% in~\autoref{sec:relevance_of_sl_task}.

\subsection{Fruit Anomaly Detection Datasets}
\label{sec:datasets}

\subsubsection{Riseholme-2021}
\label{sec:dataset_riseholme}
% \todo{Change to for example Fruit Anomaly Detection Datasets}

For realistic evaluations, we first introduce Riseholme-2021, 
a new dataset of strawberry images, in which $3,520$~images 
are available with manually annotated labels, such as 
\texttt{Ripe, Unripe, Occluded,} and \texttt{Anomalous}.
This dataset was collected by deploying a commercial mobile robot \emph{Thorvald} 
outdoors on the strawberry research farm on the Riseholme 
campus of the University of Lincoln as depicted in~\autoref{fig:thorvald}. 
In particular, the robot was configured to use a side-mounted RGB~camera 
to take images of normal and anomalous strawberries at various growth 
stages under natural, variable light conditions, 
whilst navigating along the rows in polytunnels. 
Human experts then examined each image to manually crop the regions 
centered around strawberries and annotate with respective labels. 
In real applications, fruit segmentation algorithms could be employed to 
automate the extraction of fruit-centered regions, but in this work, 
we allow humans to intervene in the loop to minimise potential negative 
impacts caused by errors in segmentation process.
% \todo{We have to clarify somewhere that the system is designed to work mainly 
% with the fruit segmentation algorithms (so that is fully automated) but for this 
% work we have considered manual annotations instead.} 

More specifically, each image from \texttt{Ripe} 
(\texttt{Unripe}) contains a single ripe (unripe) strawberry, whereas   
several ones may appear overlapping one another in images of the \texttt{Occluded} class. 
Furthermore, some \texttt{Occluded} strawberries are 
observed to be covered by green stems.  
\texttt{Anomalous} cases also display single strawberries with 
anomalies, such as the presence of malformations, the lack of normal 
pigmentation, or clear signals of disease.
Example images of each category are displayed 
in~\autoref{fig:strawberry_examples_normal}-\ref{fig:strawberry_examples_anomalous}.
\begin{table}[t]
        \centering
        \begin{tabular}{|c|c|c|c|c|c|c|}
        \hline
         & \multirow{2}{*}{All} & \multicolumn{3}{|c|}{Normal} & \multirow{2}{*}{Anomalous} \\ \cline{3-5}
         & & Ripe & Unripe & Occluded & \\ \hline\hline
         \# of Images & $3,520$ & $462$ & $2,406$ & $499$ & $153$ \\ \hline
         Percentage & $100.0\%$ & $13.1\%$ & $68.4\%$ & $14.2\%$ & $4.3\%$ \\ \hline
         Avg. W$\times$H & $63\times66$ & $75\times81$ & $59\times61$  & $71\times75$ & $60\times60$  \\ \hline
         Std. W$\times$H & $18\times23$ & $18\times22$ & $17\times22$  & $16\times21$ & $16\times17$ \\ \hline
        \end{tabular}
        \caption{Statistics of Riseholme-2021. In this work, 
                all subcategories except anomalous are included in normal class.}
        \label{table:data_stat}
\end{table}

\Autoref{table:data_stat} shows the basic statistics of the dataset,
in which ``normal'' categories, including  
\texttt{Ripe}, \texttt{Unripe}, and \texttt{Occluded}, have 
considerably more images ($95.7\%$) than the \texttt{Anomalous}
($4.3\%$)---i.e.,~this severe class imbalance provides a realistic 
testbed for anomaly detection.
Riseholme-2021 is also presented with exclusive data 
sets---\texttt{Train}, \texttt{Val}, and \texttt{Test}---which contain 
$70\%$, $10\%$, and $20\%$~``normal'' images, respectively, and 
all \texttt{Anomalous} images are considered only during test.
To further encourage active research in agri-technology, we publish 
our Riseholme-2021 dataset online at
\url{https://github.com/ctyeong/Riseholme-2021}.

\subsubsection{Fresh \& Stale}
\label{sec:dataset_fresh_stale}

% \todo{Perhaps first describe what the dataset contains: e.g. images of different fruit together with anomalies such as rot or damage collected in controled environment. Only then describe its characteristics and limitations.}

{Fresh \& Stale}\footnote{
https://www.kaggle.com/raghavrpotdar/fresh-and-stale-images-of-fruits-and-vegetables} dataset contains annotated fruit images of six different species collected in controlled environments, and each image is labelled with either \texttt{Fresh} or \texttt{Stale}. 
We, however, have discovered duplicate images that have been transformed 
with several methods---rotations or translations. 
We thus only keep images with unique instances of fruit, 
and as a result, the size of final dataset significantly is reduced, 
in which \texttt{Apple} is the largest class with $231$~normal
and $327$~anomalous instances. 
Moreover, we only utilise \texttt{Apple}, \texttt{Orange}, and 
\texttt{Banana} since other classes each have less than 
$50$~examples after the removal of duplicates. 
We set the split of normal data with \texttt{Train}~($40\%$), 
\texttt{Val}~($10\%$), and \texttt{Test}~($50\%$) to conduct 
tests along with images of \texttt{Stale}. 
Also, the black pixels that the pre-transformation had produced were 
removed by conversion to white to match the original background. 
% \todo{Move to the dataset section.}

\subsection{Implementation Details \& Evaluation Protocols}
\label{sec:implementation_and_protocols}

Throughout experiments, we deploy a deep-network classifier for SL which 
consists of $5$~ConvLayers followed by $2$~DenseLayers, in which the number 
of $3\times3$~convolutional filters incrementally increases 
($64, 128, 256, 512,$ and $512$) as each layer is followed by 
a BatchNorm layer and a $2\times2$ MaxPool layer, and the DenseLayers 
have $256$~and $1$~output nodes, respectively. 
Every layer uses LeakyReLU activations except the last layer with a 
sigmoid function.
Note that despite ResNet's successes in SL~\citep{KZB19, LSYP21}, 
we did not discover any benefit from using it in this work 
possibly because of the relatively small resolutions of our images.

Also, at each training iteration, every image is resized to $64\times64$
and processed with traditional augmentations before 
CH-Rand---horizontal/vertical flips and color jitter\footnote{https://pytorch.org/vision/stable/transforms.html} 
changing the brightness, contrast, saturation, and hue. 
Normalisation is then applied to set pixel values to be bounded by $[-1, 1]$. 
% after all augmentations have been applied.
% \todo{Is this some standard architecture used in similar applications or is there some way to justify this design? 
% Also wonder if it would be worth refering to the loss function again as it is a key component of the architecture.}  

As in previous works for OC~\citep{CPLP21, LSYP21, RVGDSBMK18}, the Area Under 
the Curve (AUC) of the Receiver Operating Characteristic~(ROC) is used as 
the performance indicator, and the AUC of Precision-Recall~(PR) curve 
is also reported as an additional metric considering the highly
imbalanced class distribution in Riseholme-2021
(cf.,~\autoref{table:data_stat}). % \todo{check c.f. as I think it might be cf.}
Each AUC~is the average of three individual runs to mitigate 
the random effects from CH-Rand and weight initialisations in networks. 

In fact, other representation learning frameworks~\citep{LSYP21,PP19}
suggest a certain number of training iterations to achieve their 
best performance, although in practice, such knowledge is unavailable 
in advance. 
Our approach is, however, to regularly monitor 
the ``validation accuracy'' to stop training if the mean of the last five measurements 
reaches $>.95$, or $1.5$K~epochs have passed to deploy the model with the maximum validation 
accuracy.  % \todo{iterations?}
These criteria apply to all SL-based methods in our experiments to examine 
the \emph{relevance} of their pretext tasks to the final task of anomaly detection. 
% \todo{Is this the place to highlight differences to other work which uses fixed no epochs?}
More details of hyperparameters with the code is available 
online at \url{https://github.com/ctyeong/CH-Rand}.

\subsection{Comparative Results}
\label{sec:comparative_results}
\begin{table*}[t]
        \centering
        \begin{tabular}{|c|c||C{9mm}|C{9mm}|C{10mm}|C{10mm}|C{9mm}|C{9mm}|C{9mm}|C{9mm}|C{10mm}|C{10mm}||C{10mm}|C{10mm}|}
        % \begin{tabular}{|c|c|c|c|c|c|c|c|c|c|c|c|c|c|c|c|}
        \hline
        Category & Metric            & HIST & VGG16\newline\citep{SZ14} & DCAE\newline\citep{CPLP21} & DSVDD\newline\citep{RVGDSBMK18} & IOGEN\newline\citep{CPLP21} & DOC\newline\citep{PP19} & ROT\newline\citep{GSK18} & CP\newline\citep{LSYP21} & CH-P\newline($k=1$)\newline\citep{LHS20} & CH-P \newline ($k=5$)\newline\citep{LHS20} & CH-R \newline($k=1$)\newline(Ours) & CH-R \newline ($k=5$)\newline(Ours) \\ \hline\hline
        \multirow{2}{*}{Ripe}   & ROC & $.915$\newline\tiny$\pm.000$ & $.810$\newline\tiny$\pm.000$ & $.874$\newline\tiny$\pm.005$ & $.912$\newline\tiny$\pm.006$ & $.887$\newline\tiny$\pm.044$ & $.820$\newline\tiny$\pm.000$  & $\textbf{.926}$\newline\tiny$\pm.003$ & $.911$\newline\tiny$\pm.004$ & $.918$\newline\tiny$\pm.005$ & $.922$\newline\tiny$\pm.005$   & $.920$\newline\tiny$\pm.004$          & $.922$\newline\tiny$\pm.001$\\ 
                                & PR  & $.943$\newline\tiny$\pm.000$ & $.864$\newline\tiny$\pm.000$ & $.907$\newline\tiny$\pm.003$ & $.935$\newline\tiny$\pm.022$ & $.893$\newline\tiny$\pm.057$& $.883$\newline\tiny$\pm.000$ & $\mathbf{.958}$\newline\tiny$\pm.002$  & $.941$\newline\tiny$\pm.006$ &  $.949$\newline\tiny$\pm.005$ & $.954$\newline\tiny$\pm.004$           & $.955$\newline\tiny$\pm.002$          & $.957$\newline\tiny$\pm.001$\\ \hline
        \multirow{2}{*}{Unripe} & ROC & $.753$\newline\tiny$\pm.000$ & $.621$\newline\tiny$\pm.000$ & $.783$\newline\tiny$\pm.003$ & $.810$\newline\tiny$\pm.004$ & $.664$\newline\tiny$\pm.046$ & $.698$\newline\tiny$\pm.001$ & $.819$\newline\tiny$\pm.007$ & $.860$\newline\tiny$\pm.006$ & $.874$\newline\tiny$\pm.012$ & $\mathbf{.876}$\newline\tiny$\pm.012$ & $.873$\newline\tiny$\pm.004$          &          $.870$\newline\tiny$\pm.006$\\ 
                                & PR  & $.556$\newline\tiny$\pm.000$ & $.340$\newline\tiny$\pm.000$ & $.486$\newline\tiny$\pm.003$ & $.531$\newline\tiny$\pm.010$ & $.342$\newline\tiny$\pm.039$ & $.452$\newline\tiny$\pm.000$ & $.554$\newline\tiny$\pm.023$ & $.674$\newline\tiny$\pm.028$ & $.709$\newline\tiny$\pm.026$ & $.721$\newline\tiny$\pm.026$          & $.755$\newline\tiny$\pm.005$          & $\mathbf{.765}$\newline\tiny$\pm.003$\\ \hline
        Ripe \&                 & ROC & $.700$\newline\tiny$\pm.000$ & $.623$\newline\tiny$\pm.000$ & $.752$\newline\tiny$\pm.002$ & $.727$\newline\tiny$\pm.005$ & $.679$\newline\tiny$\pm.023$   & $.642$\newline\tiny$\pm.001$ & $.772$\newline\tiny$\pm.009$ & $.780$\newline\tiny$\pm.013$ & $.826$\newline\tiny$\pm.004$ & $.829$\newline\tiny$\pm.003$          & $.834$\newline\tiny$\pm.008$          & $\mathbf{.838}$\newline\tiny$\pm.007$\\ 
        Unripe                  & PR  & $.385$\newline\tiny$\pm.000$ & $.274$\newline\tiny$\pm.000$ & $.426$\newline\tiny$\pm.003$ & $.378$\newline\tiny$\pm.012$ & $.334$\newline\tiny$\pm.018$ & $.331$\newline\tiny$\pm.000$ & $.428$\newline\tiny$\pm.018$ & $.452$\newline\tiny$\pm.026$ & $.572$\newline\tiny$\pm.007$ & $.580$\newline\tiny$\pm.008$          & $.607$\newline\tiny$\pm.012$          & $\mathbf{.615}$\newline\tiny$\pm.008$ \\ \hline
        All w/                 & ROC  & $.673$\newline\tiny$\pm.000$ & $.598$\newline\tiny$\pm.000$ & $.715$\newline\tiny$\pm.002$ & $.670$\newline\tiny$\pm.012$ & $.609$\newline\tiny$\pm.040$  & $.598$\newline\tiny$\pm.001$ & $.736$\newline\tiny$\pm.005$ & $.736$\newline\tiny$\pm.007$ & $.795$\newline\tiny$\pm.005$ & $.790$\newline\tiny$\pm.005$             & $\mathbf{.804}$\newline\tiny$\pm.014$ &          $.796$\newline\tiny$\pm.012$\\ 
        Occluded               & PR   & $.303$\newline\tiny$\pm.000$ & $.228$\newline\tiny$\pm.000$ & $.340$\newline\tiny$\pm.003$ & $.295$\newline\tiny$\pm.018$ & $.236$\newline\tiny$\pm.026$ & $.252$\newline\tiny$\pm.000$ & $.335$\newline\tiny$\pm.016$ & $.337$\newline\tiny$\pm.006$ & $.458$\newline\tiny$\pm.016$ & $.436$\newline\tiny$\pm.019$             & $\mathbf{.496}$\newline\tiny$\pm.022$ &          $.484$\newline\tiny$\pm.021$\\ \hline
        \end{tabular}
        \caption{Average AUC--ROC and AUC--PR scores on Riseholme-2021 with standard deviations from three independent runs for each model. HIST, VGG16, ROT, and CP also employ the $k$~nearest neighbor detector with the best $k\in\{1, 5, 10\}$.}
        \label{table:riseholme_result}
\end{table*}

We compare CH-Rand with related methods mentioned 
in~\autoref{sec:related_work}---DCAE~\citep{CPLP21},
DSVDD~\citep{RVGDSBMK18}, IO-GEN~\citep{CPLP21}, 
ROT~\citep{GSK18}, CP~\citep{LSYP21}, and CH-Perm~\citep{LHS20}. 
Note that for representation learning under~SL, ROT and CP inject 
structural irregularities into images, while CH-Perm augments with 
randomly shuffled channels. 
% . \todo{Perhaps would be worth listing all the acronyms here.}
A basic colour feature generator (HIST) is also considered, 
in which the number of pixels is counted within 
six unique ranges in each channel, and a representation of 
$6\times6\times6$~dimensional 
% \todo{why 216 and not 256?} 
colour histogram is produced per input combining
the channel-wise ranges.
Pretrained VGG16~\citep{SZ14} is also used to generate 
features to investigate the utility of the learnt features
on ImageNet~\citep{DDSLLF09}.
DOC~\citep{PP19} is also set up, which learns representations 
utilising an external benchmark dataset (e.g., CIFAR-$10$~\citep{KH09})

Hyperparameter searches are conducted for each baseline to offer the best results 
on the Riseholme-2021 dataset first, albeit initial configurations are set up based on 
publicly available codes, e.g.,~DCAE's performance dramatically improves 
with a smaller image size of $32\times32$.
Since official source codes are not available for CP, we have implemented it 
based on the details on their appendix in~\citep{LSYP21}.
In particular, we adopt the deep classifiers and the $k$~neighbors-based 
detector described in~\autoref{sec:methodology} on ROT, CP, and CH-Perm 
so as to focus only on the achieved representation power in comparison to ours.
The only distinction with ROT is to use four output nodes in the classifiers 
to predict four degrees of rotation---i.e.,~$0^{\circ}, 90^{\circ}, 180^{\circ},$ 
and $270^{\circ}$---pre-applied to input images. 
Similarly, HIST and VGG16 run the same detector to discern anomalies 
on their representations. 
In addition, the results obtained by the best~$k\in\{1, 5, 10\}$ are presented
except CH-Rand and CH-Perm. 
Note here that high-performing models on Riseholme-2021 are then applied to Fresh \&~Stale 
without major modifications to assess the general capacity on various 
environments in agriculture.
Lastly, CH-Rand uses representations~$g_\theta$ at~\texttt{fc6}, 
and discussions on this design are described in~\autoref{sec:ablation_study}.  

\subsubsection{Riseholme-2021}
\label{sec:riseholme}
\begin{table}[b]
        \centering
        \begin{tabular}{|c||C{8.8mm}|C{8.8mm}|C{8.8mm}|C{8.8mm}|C{8.8mm}|C{8.8mm}|}
        \hline
        % Category &DCAE & DSVDD &  HIST & CP & CH-P & CH-R  \\ \hline\hline
        % Apple & $.487$ \newline\tiny$\pm.008$ & $.529$ \newline\tiny$\pm.006$& $.873$\newline\tiny$\pm.000$ & $.763$ \newline\tiny$\pm.021$& $\mathbf{.898}$ \newline\tiny$\pm.011$ & $.892$ \newline\tiny$\pm.007$ \\ \hline
        % Orange & $.554$ \newline\tiny$\pm.003$ & $.721$\newline\tiny$\pm.020$& $.854$\newline\tiny$\pm.000$ & $.816$ \newline\tiny$\pm.039$& $.843$ \newline\tiny$\pm.010$ & $\mathbf{.906}$ \newline\tiny$\pm.004$ \\ \hline
        % Banana & $.845$ \newline\tiny$\pm.017$ & $.733$ \newline\tiny$\pm.015$& $.856$\newline\tiny$\pm.000$ & $.858$ \newline\tiny$\pm.013$& $.975$ \newline\tiny$\pm.010$ & $\mathbf{.992}$ \newline\tiny$\pm.001$ \\ \hline
        % All    & $.648 $\newline\tiny$\pm.013$ & $.526$ \newline\tiny$\pm.006$& $.846$\newline\tiny$\pm.000$ & $.827$ \newline\tiny$\pm.020$& $.831$ \newline\tiny$\pm.010$ & $\mathbf{.886}$ \newline\tiny$\pm.016$ \\ \hline
        Category & HIST &DCAE &  ROT & CP & CH-P & CH-R  \\ \hline\hline
        Apple   & $.873$\newline\tiny$\pm.000$ & $.487$ \newline\tiny$\pm.008$ & $.718$ \newline\tiny$\pm.030$ & $.763$ \newline\tiny$\pm.021$& $\mathbf{.898}$ \newline\tiny$\pm.011$ & $.892$ \newline\tiny$\pm.007$ \\ \hline
        Orange  & $.854$\newline\tiny$\pm.000$ & $.554$ \newline\tiny$\pm.003$ & $.812$ \newline\tiny$\pm.010$ & $.816$ \newline\tiny$\pm.039$& $.843$ \newline\tiny$\pm.010$ & $\mathbf{.906}$ \newline\tiny$\pm.004$ \\ \hline
        Banana  & $.856$\newline\tiny$\pm.000$ & $.845$ \newline\tiny$\pm.017$ & $.973$ \newline\tiny$\pm.003$ & $.858$ \newline\tiny$\pm.013$& $.975$ \newline\tiny$\pm.010$ & $\mathbf{.992}$ \newline\tiny$\pm.001$ \\ \hline
        All     & $.846$\newline\tiny$\pm.000$ & $.648 $\newline\tiny$\pm.013$ & $.733$ \newline\tiny$\pm.016$ & $.827$ \newline\tiny$\pm.020$& $.831$ \newline\tiny$\pm.010$ & $\mathbf{.886}$ \newline\tiny$\pm.016$ \\ \hline
        \end{tabular}
        \caption{Average AUC--ROC scores on {Fresh \& Stale} dataset. 
                Successful approaches in Riseholme-2021 are compared using their best $k\in\{1,5,10\}$ for $k$-neighbor detector.}
        \label{table:fresh_stale}
\end{table}

We test different combinations of normal classes against anomalies
as shown in~\autoref{table:riseholme_result}. 
Every method struggles more with \texttt{Unripe} than \texttt{{Ripe}}
and also with the cases where more normal types are involved since a larger 
variety of colours and shapes need to be modeled. % to keep separate from anomalous strawberries.  
In particular, the notable failure in DSVDD, HIST, IO-GEN, VGG16, and DOC indicates the 
challenge of the task with the strawberry images
as all the wild conditions are concerned. 

In overall, however, SL-based approaches, such as ROT, CP, CH-Perm, and CH-Rand,
demonstrate more robust performance across categories despite their relatively simple designs 
of data augmentation and self-supervision for learning. 
Still, significantly large drops of ROC are observed
in~ROT ($.926\rightarrow.736$)
and~CP ($.911\rightarrow.736$) compared to CH-Perm or CH-Rand
($.922\rightarrow.790$ in the worst model), as all normal subcategories 
are considered.
% the images of \texttt{Unripe} and 
% \texttt{Occluded} strawberries are all added to the normal class. 
Moreover, CP presents a $25\%$~lower PR than CH-Rand at least in 
~\texttt{Ripe\&Unripe}, and 
as \texttt{Occluded} class is also added, the margin increases up to larger than~$30\%$.
% Moreover, CH-Rand ($k=1$) presents a $34\%$~higher PR at least in~\texttt{Ripe\&Unripe}, 
% and as \texttt{Occluded} class is also added, the margin increases up to~$48\%$.
This trend supports our motivation (cf.,~\autoref{sec:ch_rand})
that representations of shapes could be less informative 
for identification of unhealthy fruits.

\Autoref{table:riseholme_result} also implies that though CH-Perm and CH-Rand 
are all trained for simply identifying unnatural colour patterns, their 
representations are not trivial features based on frequencies of various colours, 
because they obviously outperform HIST. 
In particular, CH-Rand provides considerably better results than~CH-Perm 
particularly in~PR when more complex normal sets are involved
probably because its higher randomness in augmentation can simulate more 
realistic colour anomalies. 
Furthermore, similar observations are obtained in any value of~$k$.
%  is not a dominating factor in the CH family.

% \textbf{Visual Understanding of Learnt Representations:} 
In addition, \autoref{fig:rep_viz} visualises representations in 
CH-Rand, in which the final features appear surprisingly useful for differentiation 
of anomalies though anomalous class was unavailable for explicit learning. 
In particular, \autoref{fig:rep_ripe_unripe} implies ambiguous appearances of 
anomalous samples to be represented between the ripe and unripe examples, so the 
final detector can take advantage of it.
For better understanding, some visual examples of successful and unsuccessful classification 
results are also shared online in the code repository.

\subsubsection{Fresh \& Stale}
% \todo{Use a consistent label throughout for the dataset so \emph{Fresh \& Stale}}

\Autoref{table:fresh_stale} shows that DCAE is not as 
effective as in Riseholme-2021 with highly varying ROC's in different categories, 
i.e.~$.487\sim.845$, since it easily overfits the less complex images with controlled backgrounds.

Interestingly, HIST works significantly better here than in Riseholme-2021
even outperforming ROT and CP probably taking advantage of homogeneous colour 
patterns in focal objects, and consequently, visual signals 
such as black spots on bananas are easily identified simply by colour 
frequencies. 
The failure of the two SL methods re-emphasises the lower utility of 
structural features in detecting anomalous fruits. 
% anomaly detection on fruits.

Also, CH-Perm struggles particularly with \texttt{Orange} and \texttt{All}, 
in which it loses even to HIST. 
CH-Rand, however, presents high performance 
across all fruit species.

\subsection{Relevance of SL task}
\label{sec:relevance_of_sl_task}

\begin{table}[t]
        \centering
        \begin{tabular}{|c||C{10mm}|C{10mm}|C{10mm}|C{10mm}|}
        \hline
                % & Patch & Sobel & Thrsh~$25\%$ & Thrsh~$50\%$ & Thrsh~$75\%$ & Sparse~$75\%$ \\ \hline\hline
        Dataset & ROT & CP & CH-P & CH-R\\ \hline\hline
        Riseholme-2021& $+.199$& $+.290$ & $\mathbf{+.811}$ & $+.739$    \\ \hline
        Fresh \& Stale & $-.554$ & $-.529$ & $-.244$ & $\mathbf{+.275}$  \\ \hline
        \end{tabular}
        \caption{\emph{Pearson} correlation coefficients between AUC--ROC scores and 
                validation accuracies measured during SL proxy tasks.
                ``All'' categories are considered for each dataset.}
        \label{table:correlation_val_acc_roc}
\end{table}

% Representation learning algorithms are typically designed to perform only 
% a certain number of training iterations found optimal in advance 
% for particular datasets~\citep{LSYP21,PP19}. 
% In practice, however, such prior knowledge is unavailable when the models 
% are deployed in novel environments. 
% In this work, thus, we have suggested using the validation accuracy 
% during~SL as a practical stopping criterion considering it as an estimator 
% of the performance for the following detection problem. 

% We also discuss whether the proxy tasks of~SL are relevant to 
% fruit anomaly detection. 
\Autoref{table:correlation_val_acc_roc} reveals the correlations 
between the validation accuracies during~SL and the ROC's finally 
achieved to examine relevance of each pretext task to the downstream 
task---detection of real anomalous fruit images. 
CH-Rand leads to positive correlation coefficients in all datasets, while 
others including CH-Perm have negative ones in {Fresh \&~Stale}. 
% The results are consistent across datasets to indicate significantly 
% higher relevance with CH-Rand and CH-Perm than CP. 
% CH-Rand than CP, which even has a negative correlation in {Fresh \&~Stale}.
In other words, successful training in the task of CH-Rand can \emph{ensure} 
representations for precise detection later, but continued training with other 
augmentations may rather \emph{degrade} the performance of final detector 
particularly on the {Fresh \& Stale} dataset.
Therefore, as designed in~\autoref{sec:implementation_and_protocols}, 
the validation accuracy is a useful, practical criterion for early stopping 
compared to manual searches for an optimal number of training iterations 
in other frameworks~\citep{LSYP21,PP19}.

% In contrast to other frameworks that manually searched for 
% an optimal number of training iterations as an additional 
% hyperparameter~\citep{LSYP21,PP19}, the validation accuracy can hence 
% be simply used as a practical metric for determination of early stopping 
% in CH-Rand, as designed in~\autoref{sec:implementation_and_protocols}.

% Also, the worst results from ROT and CP reconfirm the lower 
% utility of predicting structural irregularity as a pretext task 
% for building anomaly detectors for fruits.
% classifying atypical colour patterns can lead 
% to better representations to precisely distinguish anomalous fruits.
% the generated SL~task in ours can better represent the 
% fruit anomaly detection, so learning representations in it is more 
% beneficial.  

\subsection{Ablation Study}
\label{sec:ablation_study}
% \todo{My impression is that the comparative study would be best to present first as 
% it has more overall importance. the ablation study is a thorough characterisation of 
% the method so after a very positive and strong point on the methods superiority, all slightly 
% negative statements about the method or methodology (half of the models, smaller size, etc.) would have a different meaning.}

We here investigate the effects of various randomisation methods 
and hyperparameters that define our augmentation techniques.
To save computation time, we train models only on a half of 
training set of all normal classes in Riseholme-2021. 
Also, each image is resized 
to $32\times32$, and the utilised representations of~$g_\theta$ are 
always extracted at \texttt{conv5}~layer unless mentioned otherwise
% except 
% in~\autoref{sec:input_size} and~\autoref{sec:layer_selection} 
to focus on each parameter in order.
% in which those hyperparameters are also discussed. 
Moreover, $k$~is set to $1$~to only consider the nearest neighbor 
from training data to calculate the anomaly score.
\begin{table}[t]
        \begin{minipage}{\linewidth}
        \centering
        \begin{tabular}{|c||C{8mm}|C{8mm}|C{8mm}|C{8mm}|C{8mm}|C{8mm}|C{8mm}|}
        \hline
            & Patch & Sobel & Th$.25$ & Th$.50$ & Th$.75$ & Sp$.75$ & All\\ \hline\hline
        ROC & $.674$ & $.731$ & $.696$ & $.716$ & $.742$ & $.725$ & $\mathbf{.749}$ \\ \hline
        \end{tabular}
        \caption{Performance of CH-Rand on Riseholme-2021 depending on the  
                selection of pixels.}
        \label{table:pixel_selection}
        \end{minipage}\\\\

        % \begin{minipage}{\linewidth}
        % \centering
        % \begin{tabular}{|c||C{10mm}|C{10mm}|C{10mm}|C{10mm}|C{10mm}|C{10mm}|}
        % \hline
        %         % & Patch & Sobel & Thrsh~$25\%$ & Thrsh~$50\%$ & Thrsh~$75\%$ & Sparse~$75\%$ \\ \hline\hline
        %         & AGN$.10$ & AGN$.25$ & AGN$.50$ & CH-S & CH-P & CH-R \\ \hline\hline
        % ROC & $.710$ & $.715$ & $.699$ & $.710$ & $\mathbf{.755}$ & $.749$ \\ \hline
        % \end{tabular}
        % \caption{Performance of different randomisation techniques 
        %         on Riseholme-2021. All pixels are randomised.}
        % \label{table:rand_variants}
        % \end{minipage}\\\\

        % \begin{minipage}{\linewidth}
        % \centering
        % \begin{tabular}{|c||C{7.8mm}|C{7.8mm}|C{7.8mm}|C{7.8mm}|C{7.8mm}|C{10.6mm}|C{7.8mm}|}
        % \hline
        %         % & Patch & Sobel & Thrsh~$25\%$ & Thrsh~$50\%$ & Thrsh~$75\%$ & Sparse~$75\%$ \\ \hline\hline
        %                 & \multicolumn{3}{|c|}{CH-P} & \multicolumn{4}{|c|}{CH-R} \\ \cline{2-8}
        %         & $32$ & $64$ & $96$ & $32$ & $64$ & $64$ (\texttt{fc6}) & $96$ \\ \hline\hline
        % ROC & $.755$ & $.773$ & $.753$ & $.749$ & $.779$ & $\mathbf{.781}$ & $.775$ \\ \hline
        % \end{tabular}
        % \caption{Performance of CH-Perm and CH-Rand on Riseholme-2021
        %         varying the size of input image. CH-Rand is also tested with the representations 
        %         learnt at \texttt{fc6}~layer.}
        % \label{table:in_size}

        \begin{minipage}{\linewidth}
                \centering
                \begin{tabular}{|c||C{6.5mm}|C{5.4mm}|C{5.4mm}|C{5.4mm}|C{5.4mm}|C{5.4mm}|C{5.4mm}|C{6mm}|C{5.4mm}|}
                \hline
                                & CH-S & \multicolumn{4}{|c|}{CH-P} & \multicolumn{4}{|c|}{CH-R} \\ \cline{2-10}
                        & $32$ & $32$ & $64$ & $64$ \scriptsize\texttt{fc6} & $96$ & $32$ & $64$ & $64$ \scriptsize\texttt{fc6} & $96$ \\ \hline\hline
                ROC & $.710$ & $.755$ & $.773$ & $.769$ & $.753$ & $.749$ & $.779$ & $\mathbf{.781}$ & $.775$ \\ \hline
                \end{tabular}
                \caption{Performance of CH-Split, CH-Perm, and CH-Rand on Riseholme-2021 with
                        different input sizes. Utilising representations learnt at \texttt{fc6}~layer is also 
                        considered.}
        \label{table:in_size}
        \end{minipage}
\end{table}

\subsubsection{Pixel Selection}
\label{sec:pixel_selection}

Though CH-Rand is to apply randomised channels across all 
pixels of an image, we here explore the cases below where only some of
$n$~pixels are randomised. 
% the effect of applications to 
% described below, as each image has $n$~pixels.
%  images. 
% \todo{if this is the size of the region then maybe: the effects of application on local regions of $n$-pixel size.}
% Tested methods are below, and 
\Autoref{fig:augmentation_examples} visualises several methods: 
\begin{itemize}
        \item \emph{Patch}: Pixels inside a random rectangular patch~\citep{LSYP21}
        \item \emph{Sobel}: Pixels inside a large segmented region from 
                Sobel filter-based segmentation~\citep{scikit}.
        \item \emph{Th$\Delta$}: Pixels thresholded between rank~$r$ 
                and $r+\lceil n \times \Delta\rceil$ in grayscale image, where 
                $r \sim \mathcal{U}(1, n-\lceil n \times\Delta\rceil)$\footnote{Discrete uniform distribution}.
        \item \emph{SP$\Delta$}: Randomly sampled sparse $\lceil n \times\Delta\rceil$~pixels.
        \item \emph{All}: All pixels as proposed in~\autoref{sec:ch_rand}.
\end{itemize}

\Autoref{table:pixel_selection} reveals that CH-Rand
works poorly when objectness is not taken into account,   
because, Patch, which may position a patch %at an arbitrary location 
lying across multiple semantic objects, leads to the worst result. 
Similarly, as $75\%$~pixels are sparsely augmented in SP$.75$, the 
result is worse than Th$.75$, which tends to pick pixels 
on the same part of object as a result of thresholding. 
Sobel also supports this idea with its high~ROC. 

Another key observation is that CH-Rand on more pixels produces 
better results. 
For instance, Th$.\Delta$ presents the improvements as 
$\Delta$ increases, and finally when \emph{all pixels} are involved
as designed in~\autoref{sec:ch_rand}, the highest ROC 
is achieved. Thus, all pixels are considered hereafter.

\subsubsection{Randomisation Variants \& Input Size}
\label{sec:rand_variants}

We also explore the effect of different image sizes particularly comparing CH-Rand
with other channel-randomising methods such as CH-Split~\citep{LHSY17} 
and CH-Perm~\citep{LHS20} (cf.~\autoref{sec:channel_randomisation_for_sl}).  
% \begin{itemize}
%         \item \emph{CH-Split}: Channel Splitting, which copies a
%                 randomly chosen channel to others~\citep{LHSY17}.
%         \item \emph{CH-Perm}: Channel Permutation, which uses 
%                 a random permutation of channels without repetition~\citep{LHS20}.
%         \item \emph{CH-Rand}: Channel Randomisation in~\autoref{sec:ch_rand}.
% \end{itemize}
% \todo{Normal distribution as $\mathcal{N}$}

In~\autoref{table:in_size}, CH-Split leads to the lowest ROC implying that 
its three possible channel sequences may provide limited irregular patterns
to learn.
%  to learn compared to other methods. 
CH-Perm and CH-Rand each appear to work best with the size of $64\times64$, 
which is close to the average size of $63\times66$ in the dataset
(cf.,~\autoref{table:data_stat}). With that size, CH-Rand outperforms CH-Perm.

\subsubsection{Layer Selection}
\label{sec:layer_selection}

% To extract representations, we also attempt to use
% \texttt{fc6}, which is deployed immediately before the final output 
% layer. 
In~\Autoref{table:in_size}, more improvement is also discovered in CH-Rand 
with representations at \texttt{fc6} implying that the most discriminative 
representations are learnt there to offer the best features to 
the last \texttt{fc7}~layer. 
Note that we have consistently observed such a tendency with CH-Rand, 
though CH-Perm did not take any benefit. 
% , e.g.,~CH-Perm~$64$ achieved $.769$ with its~\texttt{fc6}. 

Thus, based on all these findings, the best configuration for each model 
has been adopted in~\autoref{sec:methodology} and \autoref{sec:experiments}.

\section{CONCLUSION \& FUTURE WORK}
% \todo{Check capitalisation in the heading}

We have shown the importance of learning representations of regular patterns in 
colour rather than in structure so as to reliably identify images of anomalous fruits. 
In particular, 
% We have proposed a novel data augmentation method for self-supervised 
% representation learning to effectively identify fruit anomalies in agri-robotic 
% applications.  
our CH-Rand method has demonstrated consistently accurate results of 
detection on all tested types of fruit compared to other baselines, which 
typically perform well only on some of them, whether to model structural or 
colour regularities. 
% In particular, all experimental results have supported our hypothesis that 
% learning irregularities in colour is more useful than learning of 
% atypical structural patterns for building precise fruit anomaly detectors.

In addition, unlike other methods, we have discovered the positive correlations 
between the success in the pretext task of CH-Rand and the performance of finally 
built anomaly detector.
Hence, the validation accuracy can be used as an useful criterion for early 
stopping during training in practice. 
For realistic scenarios of agricultural robots, we also have introduced a new 
image dataset, so-called \emph{Riseholme-2021}, containing $3.5$K~images of 
strawberries with various levels of maturity and normality. 
%  maturities  growth stages are contained along with 
% anomalous examples. %, and the data with all used codes are published 
% online to further encourage agri-robotics research.
% \todo{Move the second sentence to the end.}

In future work, fine-grained detections could be developed to spatially 
identify local anomalies. 
Also, we could study on a potential limitation of CH-Rand in case 
where imaged fruits contain severe structural damages in appearance. 

% We could also acquire an extended dataset to include more maturity stages including flowers 
% and apply the method to detection of diseases affecting plant leaves. 
% consider additional modalities of sensory 
% data such as texture to improve overall detection performance. 

% In future work, we could extend this framework to localise regional 
% anomalies to be applied on robot vision systems without 
% assist of object detectors. 
% Toward this goal, we could first attempt to collect a larger dataset with 
% spatial annotations and in higher resolutions. 
% \todo{What about other features such as texture, alternative colour spaces, 
% dealing with more growth categories, flowers, but also anomalies for plant disease.}

%\addtolength{\textheight}{-12cm}   % This command serves to balance the column lengths
                                  % on the last page of the document manually. It shortens
                                  % the textheight of the last page by a suitable amount.
                                  % This command does not take effect until the next page
                                  % so it should come on the page before the last. Make
                                  % sure that you do not shorten the textheight too much.

%%%%%%%%%%%%%%%%%%%%%%%%%%%%%%%%%%%%%%%%%%%%%%%%%%%%%%%%%%%%%%%%%%%%%%%%%%%%%%%%

{\small
    \ifx\usenatbib\undefined%
	\bibliographystyle{IEEEtran}%
    \else%
    \bibliographystyle{IEEEtranN}%
    \fi
	%\bibliography{IEEEabrv, IEEEexample}
	\bibliography{bib}
}

\end{document}